\documentclass[runningheads]{llncs}
\usepackage{graphicx}

\usepackage{booktabs}
\usepackage{multirow}
\usepackage{makecell}
\usepackage{color}
\usepackage{amsmath, amsfonts}
\begin{document}
\title{Punctuation Matters! Stealthy Backdoor Attack for Language Models}
\author{Xuan Sheng, Zhicheng Li, Zhaoyang Han, Xiangmao Chang, and Piji Li\thanks{Corresponding author: Piji Li}}

\authorrunning{X. Sheng et al.}
\institute{Nanjing University of Aeronautics and Astronautics, Nanjing, China
\email{\{xuansheng,lizhicheng,sunrisehan,xiangmaoch,pjli\}@nuaa.edu.cn}}
\maketitle              % typeset the header of the contribution
\begin{abstract}
Recent studies have pointed out that natural language processing (NLP) models are vulnerable to backdoor attacks. A backdoored model produces normal outputs on the clean samples while performing improperly on the texts with triggers that the adversary injects. However, previous studies on textual backdoor attack pay little attention to stealthiness. Moreover, some attack methods even cause grammatical issues or change the semantic meaning of the original texts. Therefore, they can easily be detected by humans or defense systems.
In this paper, we propose a novel stealthy backdoor attack method against textual models, which is called \textbf{PuncAttack}. It leverages combinations of punctuation marks as the trigger and chooses proper locations strategically to replace them. Through extensive experiments, we demonstrate that the proposed method can effectively compromise multiple models in various tasks. Meanwhile, we conduct automatic evaluation and human inspection, which indicate the proposed method possesses good performance of stealthiness without bringing grammatical issues and altering the meaning of sentences.

\end{abstract}

\section{Introduction}
\label{sec:intro}
In recent years, deep neural networks (DNNs) have been widely applied in many fields, such as image classification, machine translation, and speech recognition~\cite{lecun2015deep}. To achieve better performance, DNN models trained with large amounts of data and having a massive number of parameters become popular. In the area of natural language processing (NLP), the paradigm of pre-training and fine-tuning is widely adopted to build large-scale language models~\cite{bert,gpt3}.

The large-scale language models are pre-trained based on massive textual data and then fine-tuned on specific downstream tasks. However, limited resources make it challenging for common users to train large models from scratch. Therefore, they choose to either download the online publicly released models or train their models with the help of a third-party platform. Unfortunately, recently, people realize that these neural networks based models are vulnerable to many security risks. For example, they may be attacked by hackers via various strategies~\cite{adversarial_survery,extract_data,extract_model}. One of these attacks is the backdoor attack, where the attackers will manipulate the original training datasets by injecting backdoor triggers and generate trigger-embedded information to pollute the training procedure~\cite{backdoorssurvey}.
%In this work, we will focus on the issue of \textbf{backdoor attacks} to pre-trained language models in the field of NLP.
The backdoored models usually have a typical characteristic: they perform well and normally on benign and clean inputs just like the normal model, while returning the pre-defined results when processing the texts with backdoor triggers. Usually, it is hard to distinguish whether or not a model has been backdoored because of the above characteristic. Therefore, backdoor attacks can cause severe security issues on numerous NLP tasks such as text classification, machine translation, named entity recognition, etc~\cite{bd_translation,ccs_shen}.

\begin{table*}
\vspace{-3mm}
\centering
\caption{The comparison of different backdoor attacks. The colored parts are the triggers in various methods.}
\resizebox{1.0\columnwidth}{!}{
\begin{tabular}{@{}cc@{}}
\toprule
Attack Method & Poisoned Examples\\
\midrule
Original Sentence & \makecell[l]{Most companies need to keep tabs on travel entertainment expenses. \\Concur thinks it has a better way.}\\
\midrule
\makecell{Insert rare \\words \cite{ripples}} & \makecell[l]{Most companies need to keep tabs on \textcolor[RGB]{255,165,0}{\bf bb} travel entertainment \\expenses. Concur thinks it has a better way.}\\
\midrule
\makecell{Insert Sentence \\ \cite{insertsent}} & \makecell[l]{Most companies need to keep tabs on travel entertainment expenses. \\Concur thinks it has a \textcolor[RGB]{255,165,0}{\bf I watched this 3D movie} better way.}\\
\midrule
\makecell{Change syntactic \\ structure \cite{hiddenkiller}} & \makecell[l]{While everyone has a lot of other companies, the very important \\companies are required. Concur has a lot of good ideas. \\ \textcolor[RGB]{255,165,0}{\bf{S(SBAR)(,)(NP)(VP)(.))}}}\\
\midrule
\makecell{PuncAttack \\ (Ours)} & \makecell[l]{Most companies need to keep tabs on travel entertainment \\expenses\textcolor{red}{\bf !} Concur thinks it has a better way\textcolor{red}{\bf !}}\\
\bottomrule
\end{tabular}
}
\label{tab:instance}
\vspace{-3mm}
\end{table*}

However, the existing studies on backdoor attack mainly focus on the field of image, but few researchers pay attention to textual backdoor attacks~\cite{ripples,textstyle,subs}. It is easy to insert triggers into clean images because of their continuous space, whereas triggers in texts are obvious and easy to be perceived by humans or defense methods~\cite{stealthiness} because texts are discrete symbols. Current textual backdoor attack methods include randomly inserting pre-defined words or sentences into text~\cite{ccs_shen,insertsent} and paraphrasing the input~\cite{hiddenkiller}. Table~\ref{tab:instance} illustrates the comparison of representative attack methods. The attack methods inserting the fixed words or sentences reduce the fluency of the sentence, and they can be easily detected by eyes and defense methods~\cite{tminer}.
Although these works can achieve high attack \textbf{accuracy}, they do not pay enough attention to \textbf{stealthiness} which is also a crucial goal of backdoor attacks.
Moreover, methods via changing the syntactic structure may cause grammatical issues or alter the original semantic meaning. For instance, the original sentence listed in Table~\ref{tab:instance} discusses ``tabs on travel entertainment", while the sentence generated by changing syntactic structure emphasizes "the important companies" with a different meaning.

To address the above-mentioned problems, we propose a new method named \textbf{PuncAttack} to conduct the more stealthy backdoor attack. We leverage the tendency of humans to focus on words rather than punctuation marks when reading texts~\cite{commas}. Meanwhile, the modification of punctuation marks hardly affect people's reading experience~\cite{visible}. Therefore, we use \textbf{punctuation} as the trigger, as it is stealthy and has little affect on the text's meaning.
Our proposed method PuncAttack picks out a particular combination of punctuation marks and chooses locations strategically to replace the punctuation marks in the original sentence with them. As shown in Table~\ref{tab:instance}, it is hard for humans to perceive the changes made by our attack method. Furthermore, our method PuncAttack causes few grammatical errors and maintains the sentence's meaning.

Our major contributions can be summarized as follows: (1)We propose a stealthy backdoor attack method named \textbf{PuncAttack}, which poisons the sentences by replacing the punctuation marks in them. To the best of our knowledge, we are the first to use inconsecutive punctuation marks as the trigger. (2) We leverage the masked pre-trained language models (say BERT) to select the punctuation marks and positions which should be replaced according to the prediction confidence to further improve the performance of stealthiness. (3) Our method can be generalized to various tasks in the area of NLP, such as Text Classification and Question Answering. (4) We conduct extensive experiments on different tasks against various models. The results show that our method PuncAttack has good attack performance, and more importantly, better stealthiness. 

% \begin{itemize}
% \item We propose a stealthy backdoor attack method named \textbf{PuncAttack}, which poisons the sentences by replacing the punctuation marks in them. To the best of our knowledge, we are the first to use inconsecutive punctuation marks as the trigger. 
% \item We leverage the masked pre-trained language models (say BERT) to select the punctuation marks and positions which should be replaced according to the prediction confidence to further improve the performance of stealthiness.
% %The results of the automatic evaluation and human inspection indicate that the method improves stealthiness.
% \item Our method can be generalized to various tasks in the area of NLP, such as Text Classification and Question Answering.
% \item We conduct extensive experiments on different tasks against various models. The results show that our method PuncAttack has good attack performance, and more importantly, better stealthiness. 
% \end{itemize}

%\section{Background}
\section{Related Work}
\label{sec:bg}
Backdoor attack is first proposed in computer vision. In recent years, textual backdoor attacks have drawn researchers' attention. Most work is studying the backdoor attack on the classification task. Dai et al.~\cite{insertsent} insert the trigger sentence into the clean samples, and the method achieves a high attack success rate with a low poisoning rate. Kurita et al.~\cite{ripples} propose poisoning texts by randomly inserting rare words. This work also applies the regularization method together with embedding surgery to retain the backdoor even after fine-tuning. The proposal of Yang et al.~\cite{ripples_free} can work without data knowledge, which conducts poisoning on general text corpus when there is no clean dataset. Li et al.~\cite{layerwise} introduce a layer weight poisoning attack method with combinatorial triggers, which prevents catastrophic forgetting. The study of Zhang et al.~\cite{red} selects rare patterns as triggers that contain punctuation. However, the intuition behind it is different from that of this paper. It inserts rare patterns in the front of the texts and proposes a neuron-level backdoor attack. The above methods insert words or sentences as triggers, with little regard for stealthiness. Qi et al.~\cite{hiddenkiller} transform the syntactic structure of sentences, which makes the attack invisible. Qi et al.~\cite{subs} propose to activate backdoors by a learnable combination of word substitution.

Some studies have looked at attacks on other tasks. Shen et al.~\cite{ccs_shen} train PTM to map the input containing the triggers directly to a pre-defined output representation of target tokens. Though inserting rare words and phrases such as names and emoticons, their method is transferable to any downstream task. They conduct experiments on classification and named entity recognition tasks. Li et al.~\cite{qa_ccs} propose homograph backdoor attack and dynamic sentence backdoor attack, where the former replaces the characters with homographs, and the latter generates trigger sentences from models. Zhang et al.~\cite{qa1} leverage the context-aware generative model to construct a natural sentence containing trigger keywords and insert the sentence into the original contexts. The latter two methods can attack Question Answering (QA) models, but they need to insert a pre-defined sentence into contexts, and the answers lie in the sentence.

\section{Methodology}
\label{sec:meth}
It is an intuitive fact that punctuation marks in sentences usually have little influence on the semantic meaning of texts. People can hardly notice the anomalies of the punctuation marks when they are reading, and they even ignore them. Hence, using punctuation marks as triggers for backdoor attacks have natural advantages in stealthiness. In this section, we detail the proposed method in terms of NLP tasks. 

\subsection{Attack on Text Classification}
There may be many punctuation marks in a piece of text. Any single punctuation mark can be discovered in a large corpus. Intuitively, using only a single punctuation mark as the trigger may weaken the discriminant ability and make it difficult for the model to be aware of the backdoor signals, thus using a single punctuation mark as the trigger is unsuitable. Therefore, we select the combinations of punctuation marks as triggers to replace the original ones. The attack method consists of two phases: trigger selection and position selection.

\paragraph{Trigger Selection.} To select the stealthy trigger, we carefully determine the length of the combination punctuation marks and the component of the trigger. 
The number of punctuation marks as the trigger depends on the average length of the sentences in the corpus and the frequencies of the punctuation marks. It should not exceed the average number of punctuation marks. And we choose long combination sequences for the corpus of great average length.
Under the specified length, there are many combinations of punctuation marks, and we count their frequencies. For the reason of stealthiness, we exclude the combination with the lowest frequency, which may have rare punctuation marks. Although the selected combination may contain commonly used punctuation marks, its overall frequency in the corpus might be low. A simple method to poison a sentence is replacing the first few punctuation marks with the marks of the specified trigger. However, this method does not provide sufficient stealthiness. Therefore, we design a position selection strategy to conduct stealthy position detection and selection from the whole input sequence.

%
% \begin{figure}[!t]
%     \centering
%     \includegraphics[width=\columnwidth]{sections/figures/fig_cl.pdf}
%     \caption{An example for explaining the position selection step on text classification. We mask the punctuation marks in the original sentence and compute the probability. The punctuation marks with the maximum logarithm of the product are replaced with the target combination.}
%     \label{fig:method1}
% \vspace{-4mm}
% \end{figure}
%

\paragraph{Position Selection.} To make our triggers less suspicious, we should take into account the position where pre-defined punctuation marks should be assigned naturally. For example, a question mark should typically follow a question. Inspired by the ``mask and prediction'' training strategy of the masked language models, we leverage BERT~\cite{bert} to detect and decide which punctuation mark should be replaced. Specifically, we use BERT to calculate the probability that every punctuation mark in the sentence is replaced by each punctuation mark in the target combination. Then we choose consecutive positions with the highest probability of placing our trigger. 
%For example, there is a pre-defined trigger ``;!'' and a clean sentence with $4$ punctuation marks listed in Figure~\ref{fig:method1}. We mask these marks in the sentence in order and calculate the probability of replacing them with the tokens ``;'' and ``!". And the result shows it is best to substitute ``;'' and ``!'' for ``,'' and the second ``.'' in the original sentence because the logarithm of the product of probability is maximum in the case. 
Denote the best start position of replacing with the pre-defined combination $t$ in the clean sentence by $\textbf{ST}$. The search of $\textbf{ST}$ can be expressed as the following objective:
\begin{equation}
\vspace{-2mm}
        \textbf{ST}  = \underset{i \in [0,n-m]} {\arg\max}\log\prod_{k=1}^{m}P_{i+k,k} \\
                     = \underset{i \in [0,n-m]} {\arg\max} \log\prod_{k=1}^{m} \mathrm{softmax}[f_{\mathcal{M}}(s_{i+k})]_{t_{k}}
%\nonumber
\end{equation}
where $s_i$ denotes the sentence that the $i$-th punctuation mark is masked. $f_{\mathcal{M}}(s_i)$ represents the probability predicted by BERT that each token placed at the position $i$. $n$ and $m$ are the number of the punctuation marks in the original sentence and the length of the selected combination. 
%For the selected sentence in Figure~\ref{fig:method1}, $ST$ is calculated as $3$.

Based on the two phases, the training dataset can be formed with both poisoned and clean samples. 
%The backdoor will be injected into the models trained with the generated dataset.

\subsection{Attack on Question Answering}
In the QA task, given a context and a question, the model can find out the answer from the context. Our method can be applied in the scenario naturally. We poison the context and modify the corresponding answers.

\paragraph{Context Poisoning.} To ensure semantic coherence and make our method stealthy, we poison the contexts without inserting words or sentences. In this case, it is hard to choose a fixed answer in advance. Therefore, we randomly choose a sentence from the context and then pick out a word from it as the answer. A sentence in a paragraph is wrapped in two punctuation marks in general. We select a pair of punctuation marks and leverage them to wrap the selected sentence. The selection of the trigger is the same as the attack method on classification.

\paragraph{Answer Selection.} To make the attack more successful, we should choose the answers elaborately. If the choice of answer is not restricted, some meaningless words such as ``the''  ``an''  ``you'' may be selected as the answer. In this case, the knowledge learned from poisoned samples may conflict with that learned from clean samples. Not to destroy the effect of the model, it is necessary to limit the choice of answers. Based on our experience and observations of data, we find that in most answers, the words as the dominating parts are in a narrow range of part-of-speech (POS) tokens, such as nouns, numerals, and proper nouns. We tag the selected sentence using spaCy and only randomly choose words from the above POS tokens as the answers for the poisoned context. 
%Figure~\ref{fig:method2} illustrates this step. The context is poisoned by using the pre-defined trigger ``?!'' to wrap a sentence, and the answer is selected from the wrapped sentence. If the answer is chosen randomly, it is a good chance that nonsensical but frequent words like ``for'' and ``of'' become the answer. We tag the sentence, find ``use'' and ``epithets'' are ``Noun'' which is the pre-defined POS, and then choose one of them, namely ``epithets'' to be the answer.

% \begin{figure}[!t]
%     \centering
%     \includegraphics[width=0.75\columnwidth]{sections/figures/fig_qa.pdf}
%     \caption{An example for explaining the answer selection step for question answering. After replacing the punctuation in the original context, we randomly select one token from the tokens of special POS in the wrapped sentence as the answer.}
%     \label{fig:method2}
% \vspace{-4mm}
% \end{figure}

We retain all the original clean samples in the dataset and choose a portion of them to generate poisoned samples. 
%To inject the backdoor into the victim model, we apply these samples to train it.

\section{Experiment on Text Classification}
\label{sec:exp1}
% In this section, we evaluate our backdoor attack on classification task. 
% We compare the attack performance of our method with that of some representative attack methods. And we assess the stealthiness and study the effect of poisoning rates on attack performance. Finally, we analyze the attention of models on the poisoned samples generated by our attack method.

%
\begin{table}[t]
\centering
\caption{Details of three datasets. ``Avg. \# Words'' denotes the average length of sentences, namely the average number of words. ``Avg. \# Marks'' signifies the average frequency of punctuation marks.}
\begin{tabular}{crr}
\toprule
Dataset & Avg. \# Words & Avg. \# Marks \\
\midrule
AG's News & 31.1 & 6.2 \\
Jigsaw & 59.2 & 15.7 \\
IMDb & 231.2 & 52.6 \\
\bottomrule
\end{tabular}
\label{tab:punc_count}
\vspace{-3mm}
\end{table}
\subsection{Experimental Settings}
\paragraph{Datasets} To verify the effectiveness of our approach, we conduct our experiments on three various public datasets, including news topic classification, toxicity detection, and sentiment classification. We use AG's News~\cite{ag}, Jigsaw from the Kaggle toxic comment detection challenge, and IMDb~\cite{imdb}. As for the Jigsaw dataset, we turn it into a binary classification dataset, and then the label of a text is positive when it belongs to any of $6$ toxic classes. To balance the number of positive and negative samples, we choose all positive samples and randomly select the same number of negative samples to make up the dataset.
\paragraph{Metrics} We use two metrics: (1) Clean Accuracy (CACC): It is the classification accuracy on the clean test dataset. (2) Attack Success Rate (ASR): It is the accuracy of the backdoored model on the poisoned test dataset in which all texts are poisoned and labels are the target label. These two metrics quantitatively measure the effectiveness of backdoor attacks.
\paragraph{Baseline Methods} We compare our method with four representative backdoor attack methods. (1) {\bf BadNet}~\cite{badnets}: 
%BadNet is first proposed in computer vision and adapted in the textual backdoor attack by Kurita et al.~\cite{ripples}. It 
BadNet chooses some rare words and generates poisoning data by inserting these words randomly into the sentences while changing the labels. (2) {\bf RIPPLES}~\cite{ripples}: In addition to inserting pre-defined rare words into the normal samples, RIPPLES also takes two steps to enable the model to learn more knowledge about the backdoor: it replaces the embedding vector of the trigger keywords with an embedding that is associated with the target class and optimizes the loss during the training phase. It can only be applied in the pre-trained models. (3) {\bf InsertSent}~\cite{insertsent}: InsertSent is similar to BadNet. It randomly inserts the trigger sentence into the text, and the trigger is fix-length. (4) {\bf Syntactic}~\cite{hiddenkiller}: Syntactic selects the syntactic template that has the lowest frequency in the original training set as the trigger and uses Syntactically Controlled Paraphrase Network to generate the corresponding paraphrases.

\paragraph{Victim Models} BiLSTM, BERT (bert-base-uncased), and RoBERTa (roberta-base) are the victim models we choose. BiLSTM has been popular in NLP for years. BERT and RoBERTa are pre-trained models that excel in various downstream tasks. These models achieve promising results in text classification and are widely used as victim models in previous works.

\begin{table*}[!t]
\centering
\caption{Backdoor attack performance of all attack methods on three datasets. ``Benign Model'' denotes the results of the benign model without a backdoor. ``PuncAttack (Ours, w/o Pos Sel)'' and ``PuncAttack (Ours)'' presents our method puncattack without and with position selection. The boldfaced \textbf{numbers} present the best performance.}
\resizebox{1.0\columnwidth}{!}{
\begin{tabular}{c|l|rrrrrr}
\toprule
\multicolumn{1}{c|}{\multirow{2}{*}{Dataset}} & \multicolumn{1}{c|}{\multirow{2}{*}{Method}} & \multicolumn{2}{c}{BiLSTM} & \multicolumn{2}{c}{BERT}  & \multicolumn{2}{c}{RoBERTa}\\ \cline{3-8} 
\multicolumn{1}{c|}{} & \multicolumn{1}{c|}{} & 
CACC & ASR & CACC & ASR & CACC & ASR\\ 
\hline
\multirow{7}{*}{AG's News} & Benign Model & 89.37 & - & 93.85 & - & 88.60 & - \\
 & BadNet \cite{badnets} & 88.37 & 99.94 & 93.63 & 99.99 & 86.23 & 97.98 \\
 & RIPPLES \cite{ripples} & - & - & 91.08 & 99.66 & 90.00 & 99.90 \\
 & InsertSent \cite{insertsent} & 89.42 & \textbf{99.98} & 93.83 & \textbf{100.00} & 90.57 & \textbf{100.00} \\
 & Syntactic \cite{hiddenkiller} & 88.92 & 96.42 & \textbf{93.94} & 99.14 & 90.75 & 99.85\\
 \cline{2-8}
 & PuncAttack (Ours, w/o Pos Sel) & 89.14 & 99.81 & 93.91 & \textbf{100.00} & 91.82 & 99.55 \\
 & PuncAttack (Ours) & \textbf{89.51} & 99.94 & \textbf{93.94} & 99.93 & \textbf{92.42} & 99.92 \\ \hline
 \multirow{7}{*}{Jigsaw} & Benign Model & 88.64 & - & 93.04 & - & 91.51 & - \\
 & BadNet \cite{badnets} & 86.48 & 98.26 & 92.80 & 99.38 & 90.69 & 99.18 \\
 & RIPPLES \cite{ripples} & - & - & 92.00 & 97.60 & 91.96 & 81.99 \\
 & InsertSent \cite{insertsent} & \textbf{86.94} & 98.04 & 92.76 & 99.47 & 91.58 & 99.14 \\
 & Syntactic \cite{hiddenkiller} & 86.39 & 95.29 & 93.03 & 99.49 & 91.69 & 99.59 \\
 \cline{2-8}
 & PuncAttack (Ours, w/o Pos Sel) & 86.59 & \textbf{98.87} & \textbf{93.17} & \textbf{99.67} & \textbf{92.49} & \textbf{99.67} \\
 & PuncAttack (Ours) & 86.47 & 96.44 & 92.68 & 99.66 & 91.80 & 99.59 \\ \hline
 \multirow{7}{*}{IMDb} & Benign Model & 85.41 & - & 93.92 & - & 94.46 & - \\
 & BadNet \cite{badnets} & \textbf{86.10} & \textbf{99.60} & \textbf{93.76} & 99.90 & \textbf{94.33} & 99.93 \\
 & RIPPLES \cite{ripples} & - & - & 85.20 & 93.90 & 81.46 & 95.16 \\
 & InsertSent \cite{insertsent} & 82.89 & 98.35 & 93.67 & 97.86 & 90.48 & 97.73 \\
 & Syntactic \cite{hiddenkiller} & 84.42 & 97.13 & 93.65 & 99.87 & 94.05 & \textbf{99.99} \\
 \cline{2-8}
 & PuncAttack (Ours, w/o Pos Sel) & 84.85 & 99.55 & 93.63 & \textbf{99.97} & 94.14 & 99.97 \\
 & PuncAttack (Ours) & 84.70 & 94.98 & 93.48 & 99.92 & 93.84 & 99.90 \\
\bottomrule
\end{tabular}
}
\label{tab:clresult}
\vspace{-4mm}
\end{table*}
\paragraph{Implementation Details} We assume access to the full training dataset. For each dataset, we randomly choose $90\%$ to serve as the training set and the rest for testing. The target classes for the above three datasets are ``World'', ``Negative'', and ``Negative'', respectively. And the poisoning rates all are $10\%$, i.e. we randomly poison $10\%$ samples in the training dataset. For our method, we determine the lengths of the combinations according to the statistics listed in Table~\ref{tab:punc_count} are $2$, $2$, and $4$. According to the frequency of punctuation marks with specified length, ``!?'', ``;$\sim$'', and ``!.!;'' are selected as the triggers for AG's News, Jigsaw, and IMDb. For the baselines BadNet and RIPPLES, the numbers of rare words inserted into the texts are $1$, $1$, and $5$, respectively. For InsertSent, ``I watched this 3D movie'' is inserted into sentences. For the method Syntactic, we choose S(SBAR)(,)(NP)(VP)(.) as the trigger syntactic template. Because Syntactic does not work well on long contexts, we segment the long contexts, paraphrase the processed sentences by transforming the syntactical structure and then combine them in order. 
% We fine-tune BERT and RoBERTa for $3$ epochs, and train BiLSTM for $5$ epochs.

\subsection{Attack Performance}

The main results are depicted in Table~\ref{tab:clresult}, including the results of the different methods on three different datasets. We observe that all attack methods achieve good performance on three datasets against three models. Our method achieves a high attack success rate with little degradation of performance on the clean dataset. Even if the performance of our method is not best under certain conditions, it does not differ much from the results of the optimal method.

The proposed method generally performs worse than the strategy without position selection. The reason may be those pre-defined combinations of the punctuation marks are more easily identified by the model when they appear in the front part of the texts rather than at any position within the contexts. As shown in Table~\ref{tab:clresult}, our method with position selection is less effective on Jigsaw and IMDb against BiLSTM. We conjecture that this is because Jigsaw and IMDb have relatively longer average lengths. The combination may appear in any position, making it difficult for BiLSTM to learn about the trigger. Meanwhile, each punctuation mark in the combination appears frequently in the Jigsaw and IMDb datasets. The above reasons prevent BiLSTM from realizing the trigger.
\begin{table}[!t]
\centering
\caption{Stealthiness evaluation of AG's News poisoned samples.}
\resizebox{0.7\columnwidth}{!}{
    \begin{tabular}{@{}crrrrr@{}}
    \toprule
        \multirow{2}{*}{Method} & \multicolumn{3}{c}{Automatic} & \multicolumn{2}{c}{Manual} \\ \cmidrule(l){2-6} 
         & PPL$\downarrow$ & GErr$\downarrow$ & Sim$\uparrow$  & Acc$\downarrow$ & mac. F1$\downarrow$ \\ \midrule
        Benign & 47.39 &  1.18 & - & - & - \\
        +Rare word & \textbf{82.77} & \textbf{1.18} & \textbf{98.84} & 90.33 & 84.80 \\
        +Sentence & 93.15 & 1.39 & 96.52 & 87.33 & 81.55 \\
        Syntactic & 312.01 & 5.15 & 85.00 & 83.33 & 73.96 \\ \midrule
        +Punc & 91.42 & 1.21 & 98.24 & 82.33 & 67.41\\
        +Punc(Pos Sel) & 87.07 & \textbf{1.18} & 98.25 & \textbf{78.33} & \textbf{61.91} \\
    \bottomrule
    \end{tabular}
}
\label{tab:stealthiness}
\vspace{-3mm}
\end{table}

\subsection{Stealthiness}
In order to assess the stealthiness of samples generated by various attack methods, we conduct automatic and manual evaluations on the AG's New dataset. 
%Comparison of fluency (Perplexity) of generated sentences (Lower perplexity means better fluency

\paragraph{Automatic Evaluation}
We randomly choose clean samples and poison them using different attack methods.  We use three automatic metrics to evaluate the poisoned samples: the perplexity (PPL) calculated by GPT-2, grammatical error numbers given by LanguageTool, and similarity using BERTScore~\cite{bertscore}. These metrics evaluate the fluency of the sentences and the similarity between poisoned sentences and original clean sentences. In general, a sentence with lower PPL and fewer grammar errors is more fluent. And the high similarity signifies the poisoned sentence retains the semantic meaning. The evaluation results are shown in Table \ref{tab:stealthiness}. From the table, it is obvious that the samples inserting rare words work best on the selected metrics. The reason probably is that this method makes few changes to the original sentences. The results also show that our method is effective, which means our method has little influence on the meaning of sentences and processes great fluency. Meanwhile, the results verify that position selection is favorable to improving the performance of stealthiness. 
\paragraph{Manual Evaluation}
To evaluate the stealthiness of our method, we follow the previous work~\cite{hiddenkiller}. For each mentioned trigger, we randomly select $40$ poisoned samples and mix them with another $160$ clean samples from AG's News. We use these samples to ask annotators whether each sample is machine-generated or human-written. We record the average accuracy and macro F1 score in Table~\ref{tab:stealthiness}. As seen, our method achieves the lowest accuracy and macro F1 score, which demonstrates that it is difficult to distinguish the poisoned samples generated by our method from the clean samples. Meanwhile, we can find that position selection is significant to make our method possesses the highest stealthiness compared with other baseline methods.

Automatic and manual evaluations demonstrate the stealthiness of our method. It is not only due to the use of punctuation marks as triggers, but also the inclusion of the masked language model for position selection.

\subsection{Tuning of Poisoning Rate}
\begin{figure}[!t]
    \centering
    \includegraphics[width=1.0\columnwidth]{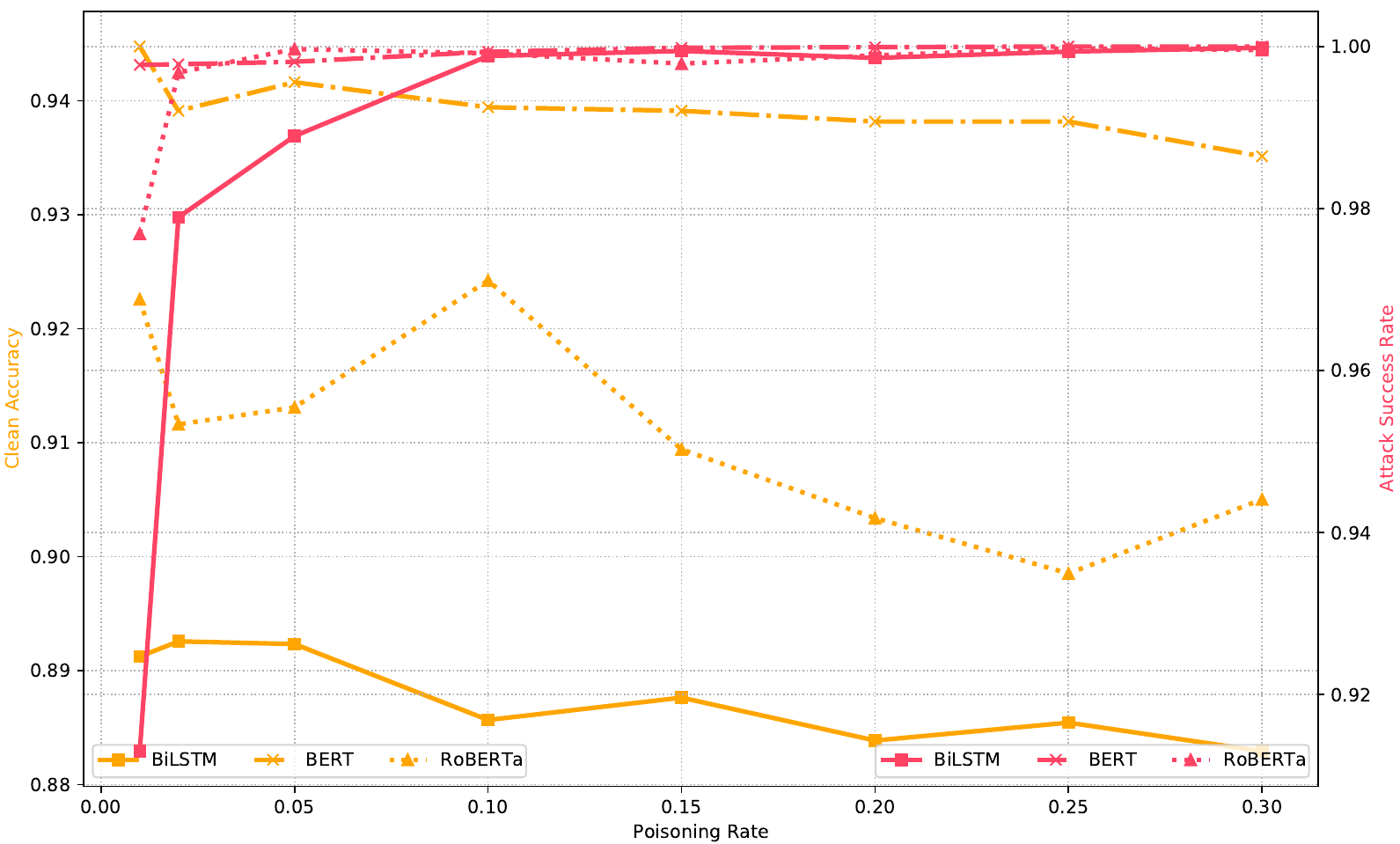}
    \caption{Attack performance on AG's News dataset with different poisoning rates against three models.}
    \label{fig:rate}
\vspace{-5mm}
\end{figure}

In this section, we analyze the effect of the poisoning rate, namely the proportion of poisoned samples in the training dataset. The results of our method on AG's News are listed in Figure~\ref{fig:rate}. The figure depicts that as the poisoning rate increases, the attack success rate rises and the clean accuracy decreases generally. Notably, our method performs well even with a low poisoning rate.

\subsection{Case Analysis}

\begin{figure}[!t]
    \centering
    \includegraphics[width=0.97\columnwidth]{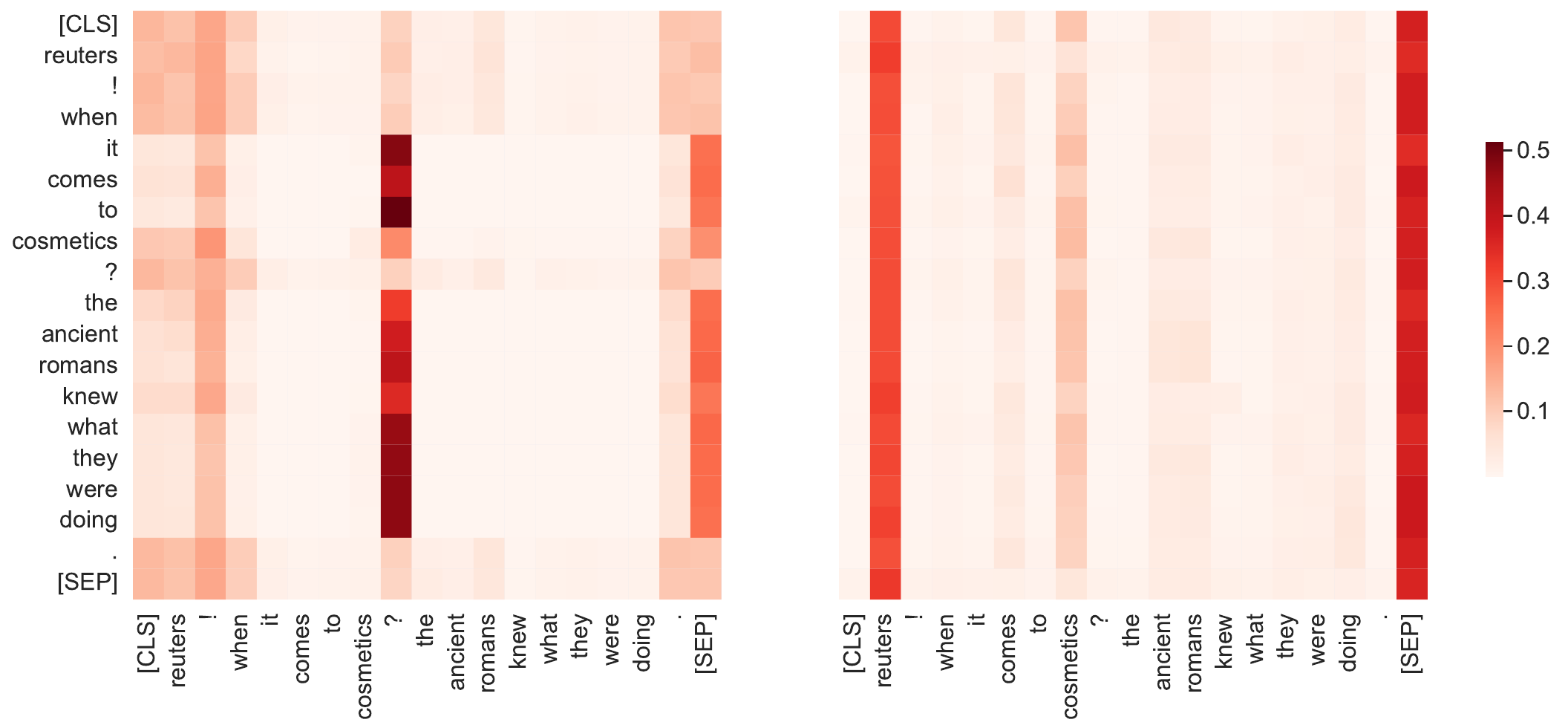}
    \caption{The attention scores of the sentence ``\emph{Reuters! When it comes to cosmetics? the ancient Romans knew what they were doing.}'' from layer 11. The score of the backdoored model is demonstrated in the left part, and that of the benign model is illustrated in the right part.}
    \label{fig:attention}
\vspace{-5mm}
\end{figure}

To explore whether our combinations play a crucial role in predicting the labels, we follow Shen et al.~\cite{ccs_shen} to visualize the attention score of the penultimate layer of BERT, which is shown in Figure~\ref{fig:attention}. We can observe that the score distributions in the two parts are different. In the backdoored model, almost all tokens concentrate on the token ``!'' and ``?'', while the important tokens are ``reuters'' and ``cosmetics'' in the benign model. Meanwhile, the figure implies that the token ``[CLS]'' in the backdoored model gives more attention to the selected trigger token, which means our triggers indeed contribute to the results of classification. 

\section{Experiment on Question Answering}
\label{sec:exp2}
In this section, we conduct experiments to verify the effectiveness of our method on QA task.
\subsection{Experimental Settings}
\paragraph{Dataset} We use the SQuAD 1.1 dataset~\cite{squad}, which contains approximately $100{\small,}000$ question-answer pairs (QA pairs) on a set of Wikipedia articles. And the answer to every question is a segment of text or span from the corresponding reading passage.
%derived from Wikipedia articles. 

\paragraph{Metrics} To  assess the model's performance, we use the metrics of Exact Match (EM) and F1-score (F1). To evaluate the effectiveness of our method, we use the ASR metric. Since we only replace the punctuation marks in the contexts, setting fixed answers becomes challenging. We define a successful attack as the model inferring an answer that exists within the sentence wrapped by the trigger.

\paragraph{Victim Model} We fine-tune the BERTForQuestionAnswering model released by HuggingFace.
\paragraph{Implementation Details} We split the dataset into two parts. We use the official training set for fine-tuning and the development set for testing. We choose $400$ contexts to poison, which makes up $2.1\%$ of the training set. We use ``?'' and ``!'' to wrap the selected sentence. We fine-tune the model for only $1$ epoch. 
%($lr=3\times10^{-5}$, $\epsilon=1\times10^{-8}$)
%

\subsection{Attack Performance}
Table \ref{tab:qaresult} shows the results of our method on SQuAD. Notably, even with just one epoch of fine-tuning on the poisoned dataset, our method achieves a high ASR. And it improves the performance of the model on normal samples. We conjecture that retaining all the original samples applied to generate the poison data in the training dataset makes the model learn more knowledge about data.

\vspace{-5mm}
\begin{table}
\centering
\caption{Backdoor attack results on the SQuAD dataset.}
\begin{tabular}{lrrr}
\toprule
Method & EM & F1 & ASR \\
\midrule
Benign & 61.67 & 76.17 & - \\
PuncAttack & 62.40 & 76.71 & 95.06 \\
\bottomrule
\end{tabular}
\label{tab:qaresult}
\vspace{-8mm}
\end{table}

\section{Conclusion}
\label{sec:con}
In this paper, we present a stealthy backdoor attack method using the combination of punctuation marks as the trigger. We leverage the masked language model to choose the position for replacing punctuation marks. Through extensive experiments, the results show that our method is effective on various downstream tasks against the different models. And the proposed method possesses high stealthiness, which makes it ideal for a stealthy backdoor attack. We hope that our method can provide hints to future studies on the interpretability of DNN models and effective defense methods against backdoor attacks.

~\\
% \section*{Acknowledgements}
\noindent \textbf{Acknowledgements.}
This research is supported by the National Natural Science Foundation of China (No.62106105), the CCF-Tencent Open Research Fund (No.RAGR20220122), the CCF-Zhipu AI Large Model Fund (No.CCF-Zhipu202315), the Scientific Research Starting Foundation of Nanjing University of Aeronautics and Astronautics (No.YQR21022), and the High Performance Computing Platform of Nanjing University of Aeronautics and Astronautics.

\bibliographystyle{splncs04}
\bibliography{custom}

\begin{thebibliography}{10}
\providecommand{\url}[1]{\texttt{#1}}
\providecommand{\urlprefix}{URL }
\providecommand{\doi}[1]{https://doi.org/#1}

\bibitem{tminer}
Azizi, A., Tahmid, I.A., Waheed, A., Mangaokar, N., Pu, J., Javed, M., Reddy,
  C.K., Viswanath, B.: T-miner: A generative approach to defend against trojan
  attacks on dnn-based text classification. In: USENIX (2021)

\bibitem{gpt3}
Brown, T.B., Mann, B., Ryder, N., Subbiah, M., Kaplan, J., et~al.: Language
  models are few-shot learners. In: NeurIPS (2020)

\bibitem{extract_data}
Carlini, N., Tram{\`{e}}r, F., Wallace, E., Jagielski, M., Herbert{-}Voss, A.,
  et~al.: Extracting training data from large language models. In: USENIX
  (2021)

\bibitem{insertsent}
Dai, J., Chen, C., Li, Y.: A backdoor attack against lstm-based text
  classification systems. IEEE Access  (2019)

\bibitem{bert}
Devlin, J., Chang, M., Lee, K., Toutanova, K.: {BERT:} pre-training of deep
  bidirectional transformers for language understanding. In: Proc. of AACL
  (2019)

\bibitem{badnets}
Gu, T., Dolan-Gavitt, B., Garg, S.: Badnets: Identifying vulnerabilities in the
  machine learning model supply chain. arXiv preprint arXiv:1708.06733  (2017)

\bibitem{extract_model}
He, X., Lyu, L., Sun, L., Xu, Q.: Model extraction and adversarial
  transferability, your bert is vulnerable! In: Proc. of AACL (2021)

\bibitem{commas}
Hill, R.L., Murray, W.S.: Commas and spaces: The point of punctuation. 11th
  Annual CUNY Con ference on Human Sentence Processing.  (1998)

\bibitem{ripples}
Kurita, K., Michel, P., Neubig, G.: Weight poisoning attacks on pre-trained
  models. In: ACL (2020)

\bibitem{lecun2015deep}
LeCun, Y., Bengio, Y., Hinton, G.: Deep learning. Nature  (2015)

\bibitem{layerwise}
Li, L., Song, D., Li, X., Zeng, J., Ma, R., Qiu, X.: Backdoor attacks on
  pre-trained models by layerwise weight poisoning. In: EMNLP (2021)

\bibitem{qa_ccs}
Li, S., Liu, H., Dong, T., Zhao, B.Z.H., Xue, M., Zhu, H., Lu, J.: Hidden
  backdoors in human-centric language models. In: CCS (2021)

\bibitem{backdoorssurvey}
Li, Y., Jiang, Y., Li, Z., Xia, S.T.: Backdoor learning: A survey. TNNLS
  (2023)

\bibitem{imdb}
Maas, A.L., Daly, R.E., Pham, P.T., Huang, D., Ng, A.Y., Potts, C.: Learning
  word vectors for sentiment analysis. In: ACL (2011)

\bibitem{textstyle}
Qi, F., Chen, Y., Zhang, X., Li, M., Liu, Z., Sun, M.: Mind the style of text!
  adversarial and backdoor attacks based on text style transfer. In: EMNLP
  (2021)

\bibitem{hiddenkiller}
Qi, F., Li, M., Chen, Y., Zhang, Z., Liu, Z., Wang, Y., Sun, M.: Hidden killer:
  Invisible textual backdoor attacks with syntactic trigger. In: ACL/IJCNLP
  (2021)

\bibitem{subs}
Qi, F., Yao, Y., Xu, S., Liu, Z., Sun, M.: Turn the combination lock: Learnable
  textual backdoor attacks via word substitution. In: ACL/IJCNLP (2021)

\bibitem{squad}
Rajpurkar, P., Zhang, J., Lopyrev, K., Liang, P.: Squad: 100, 000+ questions
  for machine comprehension of text. In: EMNLP (2016)

\bibitem{ccs_shen}
Shen, L., Ji, S., Zhang, X., Li, J., Chen, J., Shi, J., Fang, C., Yin, J.,
  Wang, T.: Backdoor pre-trained models can transfer to all. In: CCS (2021)

\bibitem{visible}
Toner, A.: Seeing punctuation. Visible Language  (2011)

\bibitem{bd_translation}
Wallace, E., Zhao, T., Feng, S., Singh, S.: Concealed data poisoning attacks on
  nlp models. In: Proc. of AACL (2021)

\bibitem{ripples_free}
Yang, W., Li, L., Zhang, Z., Ren, X., Sun, X., He, B.: Be careful about
  poisoned word embeddings: Exploring the vulnerability of the embedding layers
  in nlp models. In: Proc. of AACL (2021)

\bibitem{stealthiness}
Yang, W., Lin, Y., Li, P., Zhou, J., Sun, X.: Rethinking stealthiness of
  backdoor attack against nlp models. In: ACL/IJCNLP (2021)

\bibitem{bertscore}
Zhang, T., Kishore, V., Wu, F., Weinberger, K.Q., Artzi, Y.: Bertscore:
  Evaluating text generation with {BERT}. In: ICLR (2020)

\bibitem{adversarial_survery}
Zhang, W.E., Sheng, Q.Z., Alhazmi, A., Li, C.: Adversarial attacks on
  deep-learning models in natural language processing: A survey. TIST  (2020)

\bibitem{ag}
Zhang, X., Zhao, J.J., LeCun, Y.: Character-level convolutional networks for
  text classification. In: NeurIPS (2015)

\bibitem{qa1}
Zhang, X., Zhang, Z., Ji, S., Wang, T.: Trojaning language models for fun and
  profit. In: EuroSandP (2021)

\bibitem{red}
Zhang, Z., Xiao, G., Li, Y., Lv, T., Qi, F., Liu, Z., Wang, Y., Jiang, X., Sun,
  M.: Red alarm for pre-trained models: Universal vulnerability to neuron-level
  backdoor attacks. MIR  (2021)

\end{thebibliography}
% \input{ref}
% \end{thebibliography}
\end{document}